\documentclass[a4paper]{article}

\usepackage{INTERSPEECH2019}

\title{An Incremental Turn-Taking Model For Task-Oriented Dialog Systems}
\name{Andrei C. Coman$^1$, Koichiro Yoshino$^2$, Yukitoshi Murase$^2$, Satoshi Nakamura$^2$, Giuseppe Riccardi$^1$}
\address{
  $^1$Signals and Interactive Systems Lab, Department of Information Engineering and Computer Science, University of Trento, Italy\\
  $^2$Augmented Human Communication Laboratory, Graduate School of Information Science, Nara Institute of Science and Technology, Japan}
\email{andreicatalin.coman@studenti.unitn.it, giuseppe.riccardi@unitn.it \\\{koichiro, s-nakamura, y-murase\}@is.naist.jp}

\begin{document}

\maketitle
%
%
%
%


\begin{abstract}
In a human-machine dialog scenario, deciding the appropriate time for the machine to take the turn is an open research problem. In contrast, humans engaged in conversations are able to timely decide when to interrupt the speaker for competitive or non-competitive reasons. In state-of-the-art \textit{turn-by-turn} dialog systems the decision on the next dialog action is taken at the end of the utterance. In this paper, we propose a \textit{token-by-token} prediction of the dialog state from incremental transcriptions of the user utterance. To identify the point of maximal understanding in an ongoing utterance, we a) implement an incremental Dialog State Tracker which is updated on a token basis (iDST) b) re-label the Dialog State Tracking Challenge 2 (DSTC2) dataset and c) adapt it to the incremental turn-taking experimental scenario. The re-labeling consists of assigning a binary value to each token in the user utterance that allows to identify the appropriate point for taking the turn. Finally, we implement an incremental Turn Taking Decider (iTTD) that is trained on these new labels for the turn-taking decision. We show that the proposed model can achieve a better performance compared to a deterministic handcrafted turn-taking algorithm.




\end{abstract}


\section{Introduction}
\label{section:Introduction}

 The creation of dialog systems capable of holding conversations at the same level of naturalness as those between people is still a challenge and is far from being considered solved, in spite of the numerous studies in this field. A conversation is to be considered productive when the emphasis is given not only to the content that is conveyed, but also to the moment in which this exchange takes place. Within this paper, we address both issues by 
developing an incremental system capable of tracking the content of the conversation and identifying the appropriate moment for replying to the human counterpart.

In contrast to 
previous studies based on \textit{turn-by-turn} systems \cite{henderson2014word, smith2014comparative, sun2014sjtu, lee2014optimizing, platek2016recurrent, yoshino2016dialogue, masumura2018neural, ward2015ten}, which generate system utterances only after detecting the end of the input from the user, we place ourselves in a more challenging situation, 
where the system is expected to generate an utterance after each token coming from an ongoing user utterance, thus relying on its incremental processing \cite{zilka2015incremental, devault2009can}.

While we believe that there should be a harmonious integration between prosodic signals and lexical features \cite{atterer2008towards, raux2008optimizing}, in this work we will focus only on the utterance transcription in order to explore its potential in an incremental setting. 
Lexical features can be employed not only for determining the turn-taking point, but also as a decision point of where to start the post-processing phase, which could include dialog management and response generation.

Each interlocutor who takes part in a dialog "maintains" a so-called internal state of the dialog. This state is enriched with new information or updated during the evolution of the conversation itself. In the case of a human-machine conversation, the machine must be able to maintain a description of the human counterpart's intentions, including the grounding in the domain semantics. Consequently, an effective dialog system must be equipped with a tracker able to accumulate evidence over the sequence of utterances in the dialog and update the dialog state according to the observations. This state directly influences the behavior of the machine and its capability of identifying the point of maximal understanding of an ongoing user utterance. 
In an incremental scenario, the system can conduct post-processing and provide related responses as soon as it receives new information from the human counterpart. However, there is a trade-off between how early the turn is taken and the dialog state tracker accuracy. The latter will be reduced if the remaining part of the user utterance turns out to be informative. We thus define a new problem which consists of predicting the balance point, that trades off the accuracy reduction and the incremental processing.

We build our incremental Dialog State Tracker (iDST) by taking as reference the \textit{LecTrack} model \cite{zilka2015incremental}. The \textit{iDST} was used as a starting block for the implementation of our incremental Turn-Taking Decider (iTTD), which in turn is responsible for identifying the best point that balances the accuracy of the \textit{iDST} and the early turn-taking moment using the least amount of tokens possible. We exploit the dialog corpus annotated with dialog state released within the Dialog State Tracking Challenge 2 (DSTC2) \cite{henderson2014second} as our dialog domain. This dataset only provides the dialog state at the end of a turn and not for each token, highlighting the need for more granular feedback. To meet this need, we use the \textit{iDST} to identify the balance points within the user utterances. Those points are exploited as new labels for the pre-existing \textit{DSTC2} dataset to train the \textit{iTTD} in a supervised fashion. Source code, trained models and re-labeled dataset are available on GitHub\footnote{\texttt{https://github.com/ahclab/iDST\_iTTD}}.

\section{Task Definition}
\label{sec:task_definition}

Our final goal is to create an \textit{iTTD} (Subsection \ref{subsec:incremental_turn_taking_decider}) capable of identifying the point 
within an ongoing user utterance from where the state of the dialog relative to a specific turn no longer varies, even if new tokens 
are subsequently provided. To reach it, we need an \textit{iDST} (Subsection \ref{subsec:incremental_dialog_state_tracking}) to adapt the \textit{DSTC2} dataset to the incremental setting (Subsection \ref{subsec:dataset}).

\subsection{Incremental Dialog State Tracking}
\label{subsec:incremental_dialog_state_tracking}
The dialog state at time \(t\) can be seen as a vector \(\textbf{s}_t \in C_1 \times C_2 \times \dots \times C_k\) of \(k\) dialog state components where each component \(c_i \in C_i = \{v_1, \dots, v_{n_i}\}\) takes one of the \(n_i\) values \cite{zilka2015incremental}. The goal of a dialog state tracker consists of mapping a sequence of words \(w_1, \dots, w_n\) in a specific dialog state \(\textbf{s}_t\) at time step \(t\). This mapping is equivalent to the estimation of \(p(\textbf{s}_t|w_1, \dots, w_n)\). The latter can therefore be seen as the estimation of a joint probability over components values:
\begin{equation}
\label{equation:joint probability}
    p(\textbf{s}_t|w_1, \dots, w_n) = p(c_1, \dots, c_k |w_1, \dots, w_n, \theta)
\end{equation}
or as a product of probabilities over component values:
\begin{equation}
\label{equation:independent probability}
    p(\textbf{s}_t|w_1, \dots, w_n) = \Pi_i^k p(c_i|w_1, \dots, w_n, \theta_i)
\end{equation}
if independence between components holds. In both cases, it is necessary to determine the values of the \(\theta\) parameters.

A \textit{turn-by-turn} dialog state tracker estimates the dialog state \(p(\textbf{s}_t)\) only after processing all the tokens in the user's utterance. In contrast, a \textit{token-by-token} \textit{iDST} attempts to estimate the same dialog state \(p(\textbf{s}_t)\) after each token, thus using prefixes of the entire user utterance.

\subsection{Incremental Turn-Taking Decider}
\label{subsec:incremental_turn_taking_decider}
The dialog system must be able to determine the point when it has enough information for taking the turn. 
Therefore, this system is able to estimate, for each token in a specific turn, the following joint probability:
\begin{equation}
    p(take\_turn|\textbf{s}_t) = p(0|\textbf{s}_t) = p(c_1 = 0, \dots, c_k = 0|\textbf{s}_t, \theta)
\end{equation}
where \(\textbf{s}_t\) represents the current dialog state estimation and \(0\) indicates that there is no difference between the current dialog state estimation and the one at the end of the user's utterance. If independence between components holds, the same probability can be seen as:
\begin{equation}
\label{equation:independent probability_turn_taker}
    p(take\_turn|\textbf{s}_t) = p(0|\textbf{s}_t) = \Pi_i^k p(c_i = 0|\textbf{s}_t, \theta_i)
\end{equation}
In both cases, the complementary probability \(p(wait|\textbf{s}_t) = p(1|\textbf{s}_t) = 1 - p(take\_turn|\textbf{s}_t)\) is also estimated.

The dichotomous decision made by the \textit{iTTD} should, therefore, reflect the binary labeling that has been performed on the pre-existing \textit{DSTC2} dataset according to the dialog state estimations of the \textit{iDST}. This allows us to create a supervised version of the \textit{iTTD}. Details regarding the labeling function are provided in Subsection \ref{subsec:dataset}.

\section{Proposed Approach}
\label{sec:proposed_approach}
The \textit{iDST} consists of an encoder-based classifier. It takes as input the set of tokens \(W\), which consists of the concatenation of the system output transcript and the ASR 1-best user utterance hypothesis. Each token \(w_t \in W\) is associated with a confidence score \(a_t\)\footnote{\(a_t\) assumes a fixed value for tokens in the system's utterance. For those that are part of the user's utterance, the value assigned to the full utterance by the ASR 1-best hypothesis is used.}. To be used by the model, each token is then mapped to its corresponding fixed-size vector representation by means of the embedding function:
\begin{equation}
    \textbf{w}_t = emb(w_t)
\end{equation}
To reflect the uncertainty of the ASR, a further fully connected layer was added. This layer takes as input the concatenation of the confidence score and the previous embedding representation, thus creating a new embedding:
\begin{equation}
    \textbf{w}_t' = emb\_plus(\textbf{w}_t, a_t)
\end{equation}
This representation is then used in conjunction with the previous hidden state \(\textbf{q}_{t-1}\) by an \textit{LSTM} function, which in turn creates a new hidden state as follows:
\begin{equation}
    \textbf{q}_t = LSTM(\textbf{w}_t', \textbf{q}_{t-1})
\end{equation}
where
\begin{equation}
    \textbf{q}_{t-1} = (\textbf{c}_{t-1}, \textbf{h}_{t-1})
\end{equation}
which contains a context vector \(\textbf{c}\) and a hidden vector \(\textbf{h}\). This last vector is then used to compute, by means of a \textit{softmax} layer, the probability distribution over all the possible values that a given component can assume, which can be represented by:
\begin{equation}
\label{equation:probability over component value}
    p_t = F(\textbf{h}_t)
\end{equation}

As for the \textit{iTTD}, it must decide the ideal prefix point within the user's utterance for the dialog state prediction. In doing so, it must use as few tokens as possible while trying to maintain the same performance it would have if the entire utterance were to be used. The 
decision of whether to take the turn can be modeled as a probability distribution over two binary values by means of a \textit{softmax} layer. The same formulation used in Equation (\ref{equation:probability over component value}) can be replicated for this purpose. Function \(F\) takes as input the hidden vector \(\textbf{h}_t\) estimated previously by the \textit{iDST}, but instead of predicting the value of a component, it predicts a binary value that indicates whether the turn should be taken. Further details on data specifications are provided in Section \ref{sec:experiment}.

\section{Experiments}
\label{sec:experiment}
\subsection{Dataset}
\label{subsec:dataset}
The \textit{DSTC2} dataset operates in the restaurant information domain. The dialog state is described by means of three macro-components: \textit{Goal}, \textit{Requested} and \textit{Method}. The first macro-component is defined as the value assumed jointly by four micro-components, namely \textit{Pricerange}, \textit{Area}, \textit{Name} and \textit{Food}. 

The dataset is divided into three parts including \textit{train}, \textit{dev} and \textit{test} sets. A brief data analysis, taking into account only the user utterances, is provided in Table \ref{tab:general_information}. The \textit{train-dev} Out-Of-Vocabulary (OOV) rate is equal to \(0.21\), while the \textit{train-test} one is equal to \(0.29\). In terms of tokens distribution, all three datasets follow the Zipf's law \cite{powers1998applications} and have the same top-10 tokens set.
\begin{table}[!ht]
  \caption{Dataset analysis based on user utterances}
  \label{tab:general_information}
  \centering
  \begin{tabular}{l l l l}
    \toprule
    \multicolumn{1}{c}{} & \multicolumn{1}{c}{\textbf{Train}} & \multicolumn{1}{c}{\textbf{Dev}} & \multicolumn{1}{c}{\textbf{Test}} \\
    \midrule
    number of dialogs & $1612$ & $506$ & $1117$ \\
    number of tokens & $896$ & $720$ & $892$ \\
    max. seq. length & $28$ & $25$ & $27$ \\
    avg. tokens per turn & $3.88$ & $3.92$ & $3.67$ \\
    avg. turns per dialog & $4.93$ & $5.45$ & $5.98$ \\
    \bottomrule
  \end{tabular}
\end{table}
\begin{table}[!ht]
  \caption{iDST model layers and parameters}
  \label{tab:idst_model_parameters}
  \centering
  \begin{tabular}{ l r r }
    \toprule
    \multicolumn{1}{c}{\textbf{Layer}} & \multicolumn{1}{c}{\textbf{Parameter}} & \multicolumn{1}{c}{\textbf{Value}} \\
    \midrule
    emb & $num\_embeddings$ & $897$ \\
    & $embedding\_dim$ & $170$ \\
    \hline
    emb\_plus & $in\_features$ & $171$ \\
    & $out\_features$ & $300$ \\
    \hline
    LSTM & $input\_size$ & $300$ \\
    & $hidden\_size$ & $100$ \\
    \hline
    classifier & $in\_features$ & $100$ \\
    & $out\_features$ & component dependent \\
    & $activation$ & $log\_softmax$ or $sigmoid$ \\
    \bottomrule
  \end{tabular}
\end{table}
\begin{table}[!ht]
  \caption{iTTD model layers and parameters}
  \label{tab:ittd_model_parameters}
  \centering
  \begin{tabular}{ l r r }
    \toprule
    \multicolumn{1}{c}{\textbf{Layer}} & \multicolumn{1}{c}{\textbf{Parameter}} & \multicolumn{1}{c}{\textbf{Value}} \\
    \midrule
    & $in\_features$ & $100$ \\
    classifier & $out\_features$ & 2 \\
    & $activation$ & $log\_softmax$\\
    \bottomrule
  \end{tabular}
\end{table}
\begin{figure}[!ht]
    \centering
    \includegraphics[width=6.5cm]{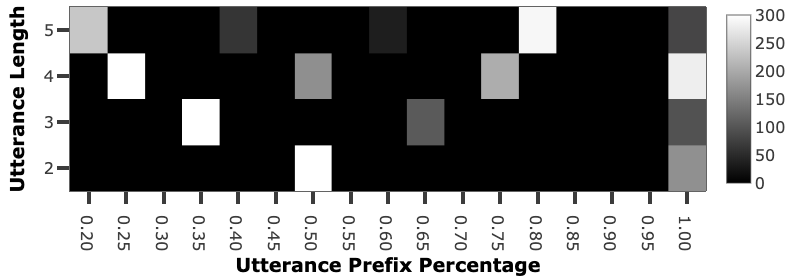}
    \caption{
    The frequency (z-axis) with which the turn is taken by the iTTD\_ASR(d = 0.85) model is computed based on the user utterance length (y-axis) and the prefix point (x-axis) of the utterance where the actual turn is taken. The frequency has been clipped to 300 for plotting enhancement.}
    \label{fig:test_percentage_utterance_length}
\end{figure}
\begin{figure}[!ht]
    \centering
    \includegraphics[width=6.5cm]{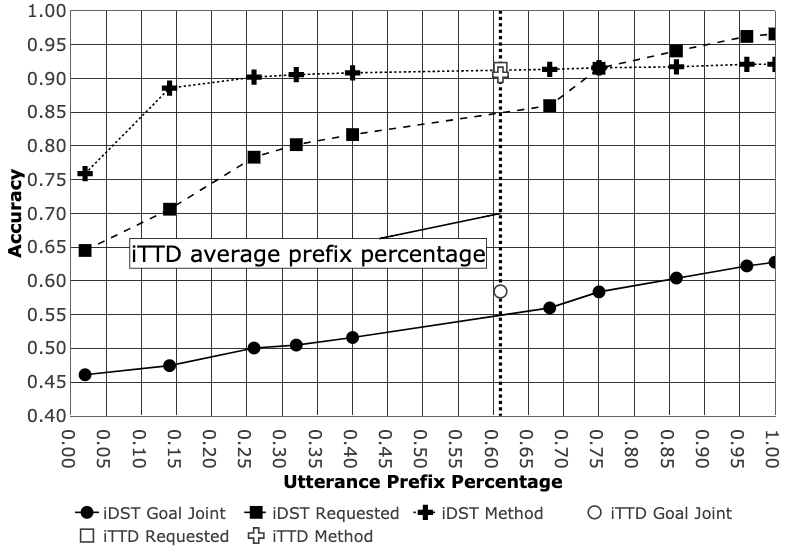}
    \caption{
    The accuracy (y-axis) of the three macro-components, namely Goal, Method and Requested, is computed in an incremental fashion based on the prefix points (x-axis) of the user utterance. The iTTD\_ASR(d = 0.85) model on average takes the turn at 61\% of the user utterance given that all micro-components in the ensemble have a confidence of at least 85\% on the predicted 0 label (turn-taking).}
    \label{fig:test_percentage_accuracy}
\end{figure}
\begin{table*}[!ht]
  \caption{
  Models with ASR suffix have been trained with the ASR 1-best user's utterance hypothesis. The ones with TRA suffix have been trained using the manual user utterance transcript. Bold indicates which one among iDST and iTTD prevailed in terms of accuracy.} 
  \label{tab:results_table}
  \centering
  \begin{tabular}{ l l l l l l l l l l l l l l l }
    \toprule
    \multicolumn{1}{c}{\textbf{}} & \multicolumn{6}{c}{\textbf{Dev}} & \multicolumn{6}{c}{\textbf{Test}} \\
    
    \multicolumn{1}{c}{\textbf{}} & \multicolumn{2}{c}{\textbf{Goal}} & \multicolumn{2}{c}{\textbf{Method}} & \multicolumn{2}{c}{\textbf{Requested}} & \multicolumn{2}{c}{\textbf{Goal}} & \multicolumn{2}{c}{\textbf{Method}} & \multicolumn{2}{c}{\textbf{Requested}} \\
    
    \multicolumn{1}{c}{\textbf{Model}} & \multicolumn{1}{c}{\textbf{Acc.}} & \multicolumn{1}{c}{\textbf{L2}} & \multicolumn{1}{c}{\textbf{Acc.}} & \multicolumn{1}{c}{\textbf{L2}} & \multicolumn{1}{c}{\textbf{Acc.}} & \multicolumn{1}{c}{\textbf{L2}} & \multicolumn{1}{c}{\textbf{Acc.}} & \multicolumn{1}{c}{\textbf{L2}} & \multicolumn{1}{c}{\textbf{Acc.}} & \multicolumn{1}{c}{\textbf{L2}} & \multicolumn{1}{c}{\textbf{Acc.}} & \multicolumn{1}{c}{\textbf{L2}} \\
    
    \midrule
    \textit{LecTrack} \cite{zilka2015incremental} & $0.63$ & $0.74$ & $0.90$ & $0.19$ & $0.96$ & $0.08$ & $0.62$ & $0.75$ & $0.92$ & $0.15$ & $0.96$ & $0.07$ \\
    \textit{iDST\_ASR(r = 1.0)} & $0.64$ & $0.53$ & $0.90$ & $0.17$ & $0.96$ & $0.07$ & $0.63$ & $0.56$ & $0.92$ & $0.13$ & $0.97$ & $0.06$ \\
    \textit{iDST\_TRA(r = 1.0)} & $0.87$ & $0.23$ & $0.94$ & $0.10$ & $0.99$ & $0.02$ & $0.82$ & $0.30$ & $0.94$ & $0.09$ & $0.99$ & $0.02$ \\
    \hline
    \textit{iDST\_ASR(r = 0.6)} & $0.57$ & $0.61$ & $\textbf{0.89}$ & $0.18$ & $0.86$ & $0.23$ & $0.56$ & $0.62$ & $0.91$ & $0.14$ & $0.86$ & $0.21$ \\
    \textit{iTTD\_ASR(d = 0.85)} & $\textbf{0.59}$ & $0.60$ & $0.88$ & $0.19$ & $\textbf{0.91}$ & $0.16$ & $\textbf{0.58}$ & $0.61$ & $0.91$ & $0.15$ & $0.91$ & $0.15$ \\
    \hline
    \textit{iDST\_TRA(r = 0.6)} & $0.77$ & $0.34$ & $\textbf{0.93}$ & $0.11$ & $0.88$ & $0.18$ & $0.73$ & $0.39$ & $\textbf{0.94}$ & $0.10$ & $0.88$ & $0.18$ \\
    \textit{iTTD\_TRA(d = 0.85)} & $\textbf{0.80}$ & $0.31$ & $0.92$ & $0.12$ & $\textbf{0.91}$ & $0.15$ & $\textbf{0.76}$ & $0.37$ & $0.93$ & $0.11$ & $\textbf{0.91}$ & $0.15$ \\
    \bottomrule
  \end{tabular}
\end{table*}
Since the dataset in question does not provide any \textit{token-level} feedback, it was necessary to find a way to propagate this information from the \textit{turn-level} to a more granular one. One of the ways in which this can be conducted is via re-labeling the dataset based on the accuracy of the dialog state prediction at the \textit{token-level}. Each token in the user utterance is assigned a binary value by means of Function (\ref{function: labeling function}). This label becomes 1 when the dialog state estimated by \textit{iDST} at \textit{i}-th token is different from the one at the last \textit{n}-th token. This also affects the accuracy value, which therefore will change. If the estimation is correct, hence the accuracy value will be the same, the label function assigns the 0 label.
\begin{equation}
\label{function: labeling function}
   label_i =
  \begin{cases}
    1 & \text{if \(Acc_i\) not equal to \(Acc_n\)}\\
    0 & \text{otherwise}
  \end{cases}
\end{equation}
Presumably, the first tokens of the user utterance will have label \(1\), while the last ones will have label \(0\). The transition point from \(1\) to \(0\) indicates a suitable moment for taking the turn. This labeling method is therefore concerned with reflecting the imposed objective, namely minimizing the number of tokens used to predict the dialog state while maintaining the same performance as if the entire utterance were to be used. These new labels, therefore, can be employed for training the \textit{iTTD}. It receives as input, for each token, the hidden vector \(\textbf{h}_t\) coming from the \textit{iDST}, and then tries to predict a binary value that reflects the labels in the new dataset, thus learning the transition point from \(1\) to \(0\). If the \textit{iTTD} confidence value on the predicted \(0\) label is greater than the imposed threshold, the \(\textbf{h}_t\) vector is then simply used by \textit{iDST} in order to estimate the dialog state.

\subsection{Experimental setting}
\label{subsec:experimental_setting}
The structure of the \textit{iDST} and \textit{iTTD} models, together with their parameters, can be found in Table \ref{tab:idst_model_parameters} and \ref{tab:ittd_model_parameters} respectively. The classifier layer in Table \ref{tab:idst_model_parameters} is a fully-connected layer and is the only one that varies according to the individual micro-components of the dialog state. 
The criterion used for the training procedure is based on the \textit{cross-entropy} loss \cite{rubinstein2013cross}. Since we are in an incremental \textit{token-by-token} setting, the value of the loss with respect to each turn is accumulated over the user tokens. This differs from a \textit{turn-by-turn} approach, where the loss value is computed only based on the prediction at the last token of the user turn. As an optimizer, we used the AMSGrad \cite{reddi2019convergence} variant of the Adam \cite{kingma2014adam} algorithm with \(learning\ rate\), \(\beta_1\), \(\beta_2\), \(eps\), and \(weight\_decay\) set to \(0.001\), \(0.9\), \(0.999\), \(1e^{-8}\) and \(0\) respectively.
The metrics used to measure the tracker performance are \textit{accuracy} and \textit{L2-norm}. The first measures the raw 1-best \textit{accuracy} of the ratio of turns in which the tracker's hypothesis is correct. The second, on the other hand, measures the \textit{L2-norm} between the distribution of scores output by the tracker and the label \cite{henderson2014second}.
\subsection{Experimental results}
\label{subsec:experimental_results}
\textit{iDST} and \textit{iTTD} implement Equation (\ref{equation:independent probability}) and (\ref{equation:independent probability_turn_taker}) respectively, by creating an ensemble of independent models, each of which refers to one of the micro-components. The results obtained by those models and a comparison with the reference one are reported in Table \ref{tab:results_table}. As a comparative analysis, we also decided to train models which, instead of using the ASR 1-best hypothesis (ASR suffix), use the manual transcription of the user utterances (TRA suffix). Variables \(r\) and \(d\) in brackets indicate the ratio of used utterance and confidence of turn-taking respectively. Since the length of the user utterances is not fixed, it was necessary to shift the imposed \(60\%\) prefix point to the nearest token, which would inevitably alter this imposed percentage. The exact ratio has therefore been computed as the average of the actual percentage values, including shifts. This implies that iDST\_ASR(r = 0.6) actually uses, as an average percentage value, \(61\%\) and \(68\%\) of the user utterance on the \textit{dev} and \textit{test} set respectively, instead of the 60\% imposed value. iDST\_\{ASR, TRA\}(r = 1.0) and iDST\_\{ASR, TRA\}(r = 0.6) use respectively 100\% and 60\% of the user utterance for dialog state estimation. iTTD\_\{ASR, TRA\}(d = 0.85) indicates that all micro-components in the ensemble must have a confidence of at least 85\% on the predicted 0 label (turn-taking). Setting the confidence threshold value to 85\% causes the iTTD to take the turn on average at 61\% (r = 0.61) of the user utterance, thus making it comparable with the deterministic iDST\_\{ASR, TRA\}(r = 0.6). The trend of the accuracy curves obtained by \textit{iDST\_ASR(r = 1.0)} relative to the three macro-components, together with the prefix point value selected on average by \textit{iTTD\_ASR(d = 0.85)} for the turn-taking decision, is shown in Figure \ref{fig:test_percentage_accuracy}. It can, therefore, be observed that the \(0.61\) ratio prefix point chosen on average by the \textit{iTTD\_ASR(d = 0.85)} obtains a better or comparable performance with respect to the deterministic iDST\_ASR(r = 0.6) which uses only \(60\%\) of the user utterance.
The \textit{iDST\_TRA} and \textit{iTTD\_TRA} models trained on the manual user transcript show how the output of the ASR negatively affects the performance of the models.
Figure \ref{fig:test_percentage_utterance_length} instead shows in greater detail the frequency with which the \textit{iTTD\_ASR(d = 0.85)} decides to take the turn based on the prefix point and the length of the user utterance. For instance, if we consider user utterances of length 2 (e.g. "phone number", "thank you", "good bye", "price range"), it can be observed that a single word, which corresponds to 50\% of the utterance, is often enough for the turn taking decision.

\section{Related Work}
\label{sec:related_work}
Capability of emulating humans behavior together with naturalness and effectiveness during a conversation, are characteristics that automatic dialogue systems must have. In this sense, several studies have been conducted such as that of \cite{chowdhury2016predicting}, which considered user satisfaction through automatic analysis of behavior by measuring emotional states and providing a description as the interaction evolves. 
The user barge-in problem was addressed by \cite{matsuyama2009enabling}, who developed a barge-in-able conversational dialog system that accepts user's barge-in utterances.
To coordinate smooth exchange for speaking turns, \cite{meena2014data} made use of prosodic, syntactic and gesture features for detecting suitable feedback response locations in the user speech. To cope with incorrectly segmented utterances, \cite{komatani2017user} proposed an a posteriori restoration methodology. To better understand the behavior of the human counterpart, \cite{DBLP:journals/corr/abs-1811-04369} tried to simulate the user by creating a model that takes into account both her initial goal and responses during the conversation. 
Despite the numerous improvements introduced, these systems have a common denominator consisting of a relatively rigid structure due to their \textit{turn-by-turn} nature. For a system to be able to replicate human behaviour during a conversation, a paradigm shift is necessary, i.e. the system must be incremental. This means that the system does not need to wait for the end of the user utterance to process it and can, therefore, perform different actions or provide feedback while listening to the human counterpart \cite{tanenhaus1995integration, schlangen2009general, dohsaka1997system, allen2001architecture}.
To improve the efficiency of the dialogue, \cite{khouzaimi2015turn} defined a turn-taking phenomenon taxonomy, and 
showed that only some phenomena are worth replicating.
ASR and NLU features have been exploited by \cite{atterer2008towards} and \cite{raux2008optimizing} in order to detect the end of the turn in an incremental setting. They showed that the combination of prosodic and lexical features can lead to promising results. A turn-taking model based on multitask learning was proposed by \cite{hara2018prediction}, which also took into account the prediction of backchannels and fillers. An incremental turn-taking model with active system barge-in was proposed by \cite{zhao2015incremental}, who modeled the turn-taking problem as a Finite State Machine and learned the turn-taking policy by means of reinforcement learning.
Our problem setting is similar to the one posed by \cite{devault2009can}, where they exploited the ASR and NLU for learning the point of maximal understanding of an ongoing user utterance. In our case, we exploit the sole ASR 1-best hypothesis and the re-labeled dataset, and try to predict the dialog state of the full utterance before it has been completed.

\section{Conclusions}
\label{sec:conclusions}
In this paper, we proposed a methodology that exploits 
lexical features to build an automated system capable of estimating the dialog state and the appropriate point for taking the turn in an incremental setting. An automatic re-labeling method that allows for the propagation of \textit{turn-level} feedback to a more granular \textit{token-level} one was introduced. Thanks to these labels we were able to create a system that, on average, performs better than a deterministic decider concerning the turn-taking problem. The decision of the threshold value regarding the confidence of the \textit{iTTD} model still remains a hyperparameter that must be manually set. This limitation opens up for future work that can focus on this problem. In addition, it is evident that there is a need for a metric capable of measuring the turn-taking performance, which is specific to the incremental setting. As future work, we would like to replace our supervised model for the decision on turn-taking with a version based on reinforcement learning. The aim is to analyze how much this methodology is able to manage the adversative relationship between minimizing the number of needed tokens and maximizing the performance.

\section{Acknowledgments}
\label{sec:acknowledgments}
The research leading to these results has received funding from the European Union --- H2020 Programme under grant agreement 826266: COADAPT.

\bibliographystyle{IEEEtran}

\bibliography{mybib}

\end{document}